\documentclass[
]{ceurart}

\usepackage{listings}
\usepackage{graphicx}
\usepackage{subcaption}
\usepackage[utf8]{inputenc} 
\usepackage[T2A]{fontenc}    
\usepackage[russian,english]{babel}
\usepackage{CJKutf8}
\usepackage{microtype} 
\usepackage{longtable} 
\usepackage{array} 
\usepackage{booktabs}
\usepackage{tikz}
\usetikzlibrary{shapes.geometric, arrows}

\tikzstyle{startstop} = [rectangle, rounded corners, minimum width=2.5cm, minimum height=1cm, text centered, draw=black, fill=red!30]
\tikzstyle{process} = [rectangle, rounded corners, minimum width=2.5cm, minimum height=1cm, text centered, draw=black, fill=orange!30]
\tikzstyle{decision} = [diamond, minimum width=2.5cm, minimum height=1cm, text centered, draw=black, fill=green!30]
\tikzstyle{arrow} = [thick,->,>=stealth]
\tikzstyle{altprocess} = [rectangle, rounded corners, minimum width=2.5cm, minimum height=1cm, text centered, draw=black, fill=blue!30]

\begin{document}

\copyrightyear{2024}
\copyrightclause{Copyright for this paper by its authors.
  Use permitted under Creative Commons License Attribution 4.0
  International (CC BY 4.0).}

\conference{CHR 2024: Computational Humanities Research Conference, December 4–6, 2024, Aarhus, Denmark}

\title{Automatic Translation Alignment Pipeline for Multilingual Digital Editions of Literary Works}

\author[1]{Maria Levchenko}[%
orcid=0000-0002-0877-7063,
email=maria.levchenko@studio.unibo.it,
url=https://mary-lev.github.io/,
]
\address{University of Bologna}

\begin{abstract}
  This paper investigates the application of translation alignment algorithms in the creation of a Multilingual Digital Edition (MDE) of Alessandro Manzoni's Italian novel \textit{I promessi sposi} (``The Betrothed''), with translations in eight languages (English, Spanish, French, German, Dutch, Polish, Russian and Chinese) from the 19th and 20th centuries. We identify key requirements for the MDE to improve both the reader experience and support for translation studies. Our research highlights the limitations of current state-of-the-art algorithms when applied to the translation of literary texts and outlines an automated pipeline for MDE creation. This pipeline transforms raw texts into web-based, side-by-side representations of original and translated texts with different rendering options. In addition, we propose new metrics for evaluating the alignment of literary translations and suggest visualization techniques for future analysis.

\end{abstract}


\maketitle
\section{Introduction}

From the very beginning of digital edition creation, there has been a tendency, supported by the power of web technologies, to represent not only the original text but also its translation(s), following the tradition of bilingual printed editions. In this paper, we propose to define multilingual digital editions (MDE) as editions in which translations are not supplementary but essential, intended to enrich both computational analysis and reader experience.

Beyond annotated file accessibility, the MDE should meet additional criteria to be effective. Primarily, the platform must display the original text alongside translations. It is anticipated that there will be a visual correlation between aligned pairs, which will facilitate straightforward comparison and analysis. The accuracy of alignment is, by default, ensuring that the corresponding parts of the texts are properly aligned. Furthermore, the platform should support the visual highlighting of omitted or inserted parts in the translations\cite{beyond}, which will enable users to discern differences and interpret the nuances of each translation.

These requirements are generally feasible for short, structured texts like poetry or historical documents (examples of MDE publishing strategies are described in Appendix A). The challenge is to develop a flexible, automated system that accurately aligns complex literary texts across multiple languages for computational analysis and user-friendly exploration. The technology should be able to handle the complexities of literary texts, including the splitting, merging, and reordering of sentences, and align text fragments of manageable length, ensuring that they are easy for users to read and understand at a glance to obtain insight into the linguistic and cultural nuances of each version. The automated alignment process should save researchers both time and resources.

For the MDE of Alessandro Manzoni's novel \textit{"I promessi sposi"} (\textit{The Betrothed}), we propose an automatic translation alignment pipeline that adapts state-of-the-art alignment techniques to the objectives of the multilingual digital edition of literary works for educational and research purposes. 

\section{The Betrothed by Alessandro Manzoni and Its Translations}
A comparative analysis of translations of the same literary work over time can provide valuable insights into the evolution of interpretation and understanding. “\textit{I promessi sposi}” is particularly compelling in this context. Not only does it reflect the author's exploration of the Italian language during a period of significant linguistic evolution, but it has also been translated into many European languages over the past two centuries. This makes it an ideal case study for investigating the influence of temporal factors, linguistic shifts, and the reception of the original novel in different cultural contexts.

Two main original editions (1827 and 1840) were translated into European languages and published in parallel in the XIX century. For the development of an automated translation alignment pipeline, we selected and prepared the texts of a wide range of translations of the classic edition of the novel, also known as Quarantana (1840), including English translations from 1845, 1876, 1983, and 2022; Russian translations from 1854 and 1999; a Dutch (1849); a German (1884), a French (1874), a Spanish (1858), a Polish (1882) and a Chinese (1998) (see \textit{Appendix 1} for a list of translations).

\section{Related Work}

The core of the MDE creation process is translation alignment, which involves mapping corresponding units (typically words or sentences) between a source and target text. State-of-the-art alignment algorithms have evolved significantly in recent years and now perform optimally in many applications, including machine translation, bilingual dictionary creation, and parallel corpus development.  

Modern methods have moved from statistical approaches \cite{church-1993-char, simard-etal-1992-using, mcenery, kraif, yet-another-fast, gale-church-1993-program} and lexical associations (Hunalign in \cite{Varga2007ParallelCF}), first to the use of machine translation (MT) systems and then to the alignment systems adopted multilingual sentence embeddings, which significantly improves the accuracy (LASER in \cite{artetxe-schwenk-2019-margin} and LaBSE in \cite{feng-etal-2022-language}). Thomson and Koehn's Vecalign \cite{thompson-koehn-2019-vecalign} uses LASER embeddings and a recursive dynamic programming approach to achieve state-of-the-art results by reducing complexity from quadratic to linear \cite{thompson-koehn-2019-vecalign}. These methods use multilingual models to generate embeddings for each sentence, which are then compared using cosine similarity to find the best matches between the original and translated sentences. Liu and Zhu \cite{10.1093/llc/fqac089} introduced Bertalign, which uses LaBSE vectors and demonstrated superior performance on the Bible One dataset and an English-Chinese literary corpus.

\section{Challenges of the Sentence-Level Alignment}

While the alignment of text and translation at the line level is sufficient for poetic, historical, and even verse dramatic texts (see \cite{alignviz}), where we cannot expect significant variation in the splitting, merging, or reordering of lines, this alignment approach is inadequate for prose due to the extent of restructuring that inevitably occurs in literary prose translation. In such cases, the standard approach is sentence-level alignment. However, it can be challenging, particularly in the case of literary translations, due to the irregularity of the syntactic structure of the original text in another language. Literary translators are not limited to translating a single sentence into another single sentence (this can be described as a one-to-one type of alignment) but are free to manage sentence boundaries and reconfigure sentence structures to better convey the meaning and style of the original text. In this case, in working to achieve the highest similarity score for the aligned pairs, the alignment algorithms are forced to combine several sentences into one, using one-to-many, many-to-one, and many-to-many alignment types.

\begin{figure}
  \centering
 \includegraphics[width=0.7\linewidth]{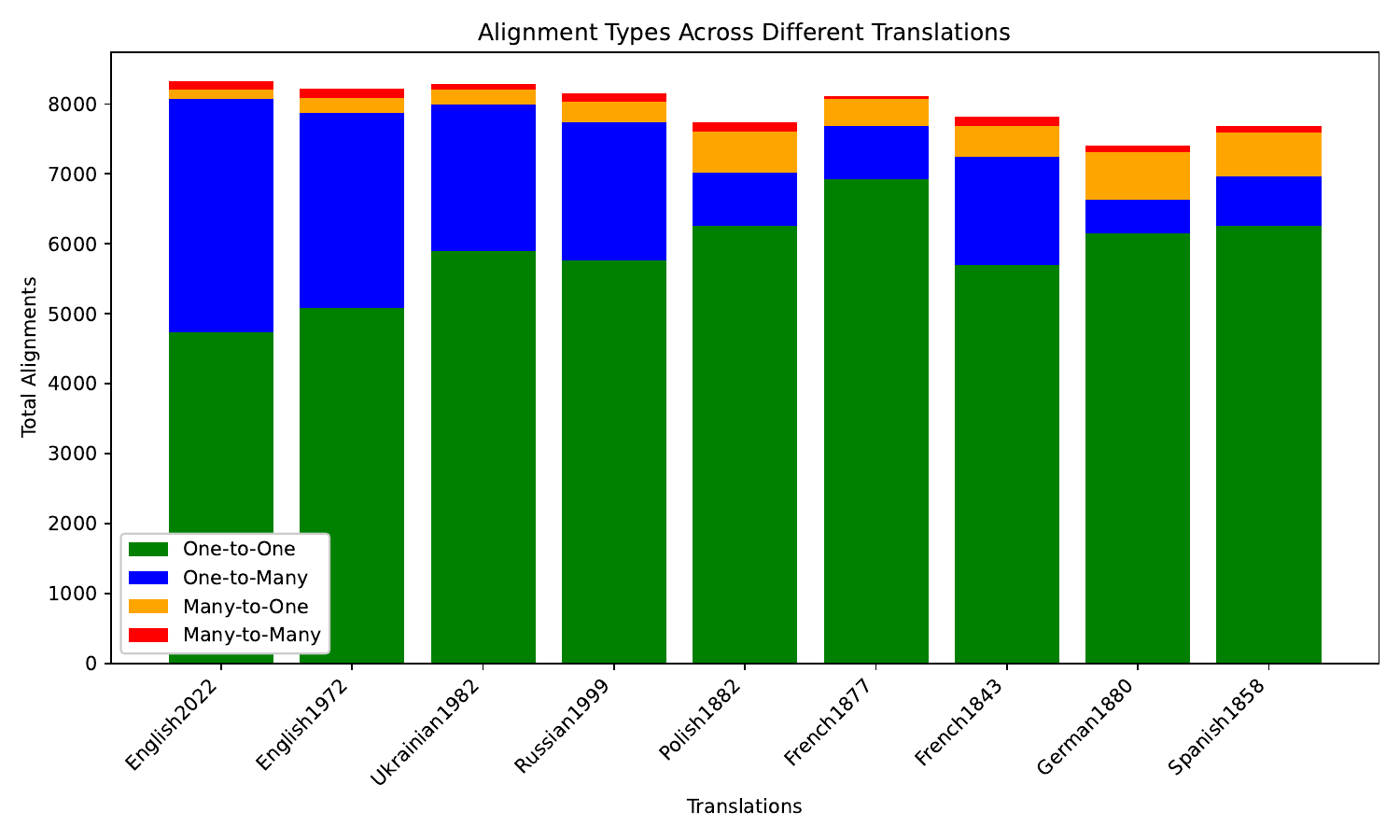}
  \caption{Alignment Types in \textit{The Betrothed}}
  \label{fig:alignmenttypes}
\end{figure}

The ideal alignment type for the MDE is a one-to-one alignment type to maintain the granularity and consistency of the alignment. In our analysis of the sentence-level alignment of \textit{I promessi sposi} (see Figure \ref{fig:alignmenttypes} for the different translations), while one-to-one alignments are the most common, a significant proportion are more complex types. While this does not inherently complicate the alignment process, as advanced tools such as Bertalign and Vecalign can handle this complexity, the results may be less optimal in terms of meaningfulness. The length of the aligned pairs becomes longer, including several sentences from both the source text and the translation (examples can be seen in the Appendix B). The edge case for this expanded alignment result would be the pairing of the paragraph or even the chapter of the original text with the same of the translation.  

That’s why traditional metrics may not be sufficient for evaluating the alignment results. The performance of alignment algorithms is typically evaluated using established metrics such as precision, recall, F1 score, and Alignment Error Rate (AER) \cite{yousef-etal-2023-evalign}. The first limitation of this approach is that it is based on a "gold dataset," which does not provide insight into the performance of the algorithm with respect to other types of text \cite{fraser-marcu-2007-squibs}. A second consequence is that the scores may be high, but the results are not suitable for MDE because the aligned pairs are too large to be analyzed or identified at a glance by a human observer. We, therefore, suggest that, in addition to the increasing importance of \textbf{the distribution of alignment types} (one-to-one, one-to-many, many-to-one, many-to-many) as a metric of the acceptability of the results, \textbf{the number and length of alignment pairs} derived from the original sentences should also be considered. A number of aligned pairs close to the number of original sentences would indicate an effective alignment process. Conversely, a significant reduction in the number of aligned pairs would indicate limitations of the sentence-level alignment approach, as it implies that the alignment algorithm is forced to combine more sentences to obtain the appropriate similarity score. 

The length of alignment pairs will indicate if they are suitable for human readers. In the context of creating a digital edition for educational or research purposes with multiple languages, it is not advisable to present long-aligned texts, given the limited attention span and working memory of the readers (for further insight, see studies of working memory and comprehension with multiple text reading \cite{comprehension, working-memory}).

\begin{figure}
  \centering
  \includegraphics[width=0.7\linewidth]{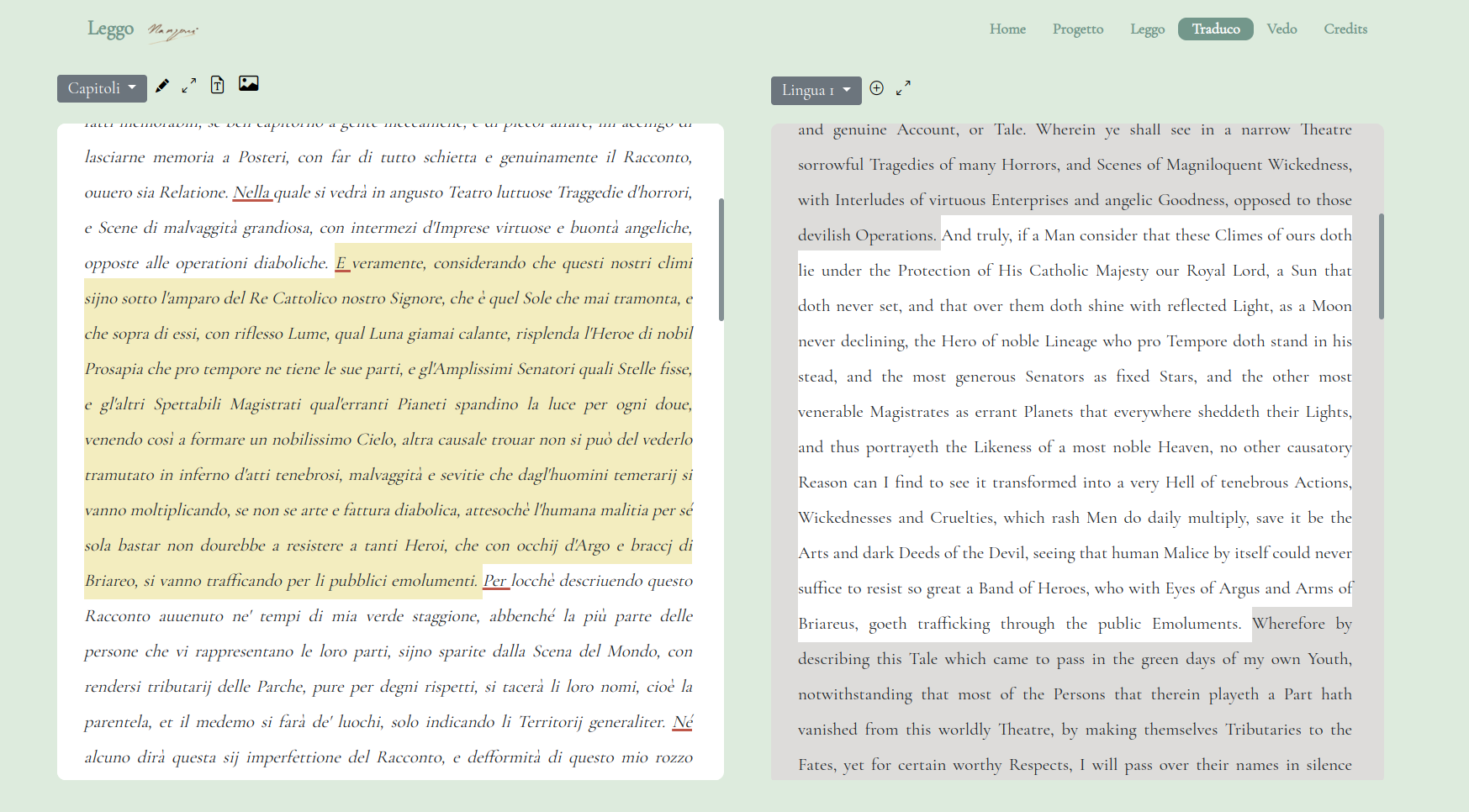}
  \caption{The visualization of the alignment of the long sentence.}
  \label{fig:sentence}
\end{figure}

To illustrate, if a sentence in the source language (Italian) is 130 tokens long and its corresponding sentence in the target language (English) is 140 tokens long, readers may encounter difficulties in comparing and understanding such lengthy segments. Even the use of color differentiation to highlight aligned pairs does not overcome this challenge (see Figure \ref{fig:sentence}).

In summary, in the context of MDE of literary works, sentence-level alignment still faces a significant challenge. 1) Sentence boundaries are not stable in different languages, which leads to a variety of alignment types and doesn't allow a consistent alignment across the MDE. 2) Strict sentence-level alignment does not fully reflect the variability of the translated texts, such as inserted or omitted parts, and 3) strains readers' attention spans and working memory and fails to achieve the alignment granularity that is comfortable for the overall reading experience. The alignment process needs to be modified to address these challenges to automated processing and readability.

\section{Sentence Segmentation as an Alternative Solution}
The alternative methods can provide more accurate and meaningful segmentation of literary texts. In an attempt to move from sentence-level alignment to phrase- or segment-level alignment, here are two promising approaches: 
\begin{itemize}
    \item Punctuation splitting:  Applying punctuation marks (such as commas, periods, and semicolons) to create initial segments. This method provides natural breaks in the text, preserving the contextual meaning. However, by using this approach and aligning the resulting segments with Bertalign, we achieved a more granular alignment but increased the number of reordering problems that didn't occur with sentence-level alignment.
    \item Zero/Few-Shot Prompting with LLM Models: The sentences of the original text are segmented using zero-shot prompting OpenAI CPT-4o model \cite{openai}. The approved segments are then used as patterns for few-shot prompting to segment the sentences of the translations. This approach provides a robust foundation for universal alignment.
\end{itemize}
The similarity score can be visualized to evaluate segment-level alignment and compare its results with traditional sentence-level alignment. In addition, the visual representation of the similarity score of the aligned segments or sentences allows us to find the semantic outliers.

\begin{figure}[htbp]
  \centering
  \includegraphics[width=0.7\linewidth]{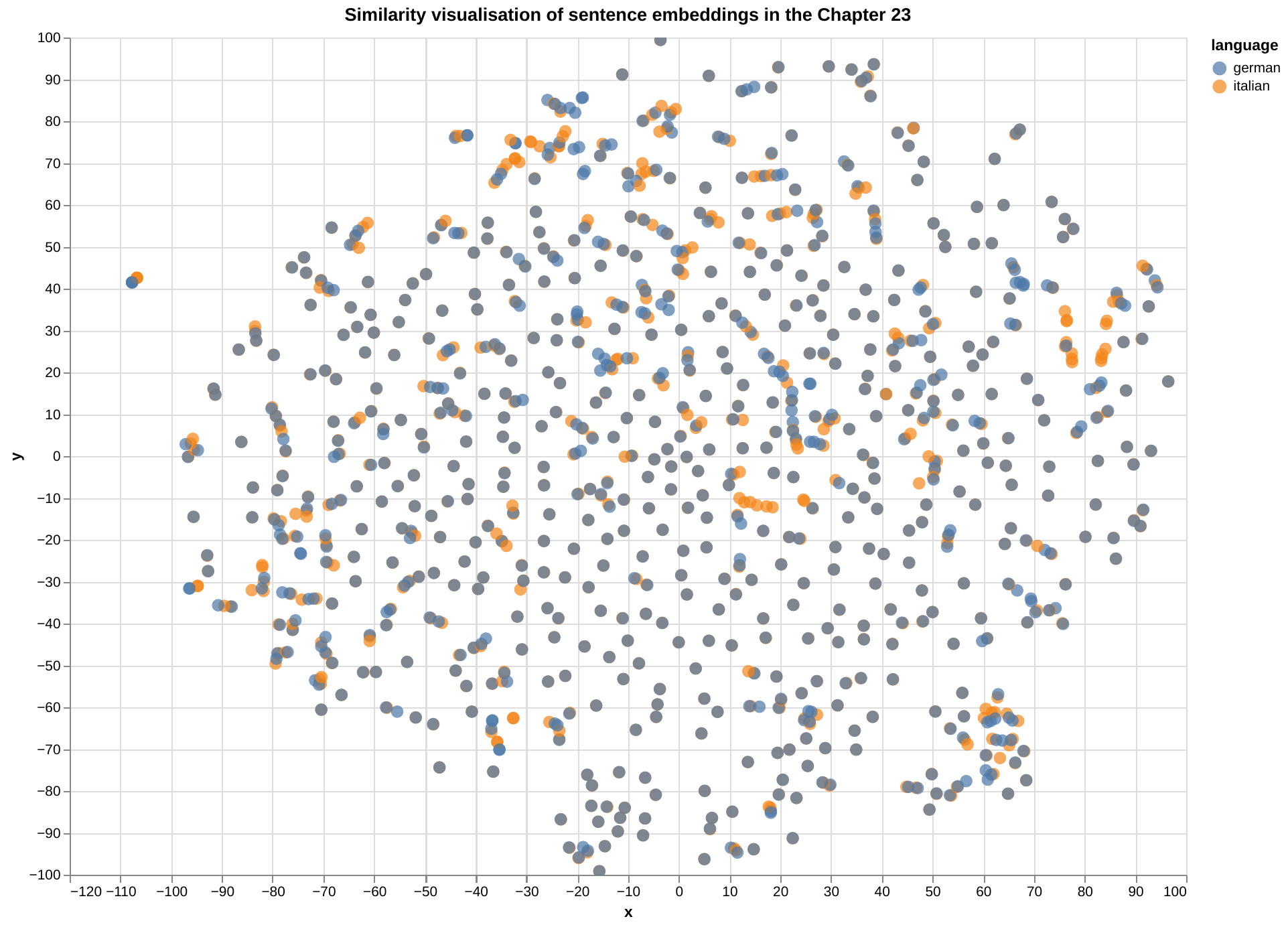}
\caption{Similarity visualisation for sentence-level alignment in the German translation of chapter 23. This translation omits Don Abbondio's inner monologue, which is not captured in the sentence-level alignment, but is evident in the visualisation, where several Italian sentences appear without corresponding German pairs.}
  \label{fig:german}
\end{figure}

After extracting high-dimensional embeddings for each aligned line from the original and translated text using the multilingual model (LaBSE), we applied t-Distributed Stochastic Neighbour Embedding (t-SNE) to reduce them to two dimensions. By visually examining the cosine similarity, we can detect anomalies and curious translated fragments (see Figure \ref{fig:german}), even if the alignment algorithm establishes the correlation between sentences.

There are several quantitative metrics that can be used to assess the quality of alignment:
\begin{itemize}
  \item \textbf{Number of resulting aligned pairs} with respect to the number of input segments;
  \item \textbf{Consistency:} Ensure that segments are consistently aligned across languages, preserving the meaning and context of the original text.
  \item \textbf{Number of clusters}: The number of clusters indicates how the sentences are grouped. An ideal alignment would result in clusters where each cluster contains two points corresponding to the embedding of the aligned pair, one from the original text and one from the translation, placed close to each other;
  \item \textbf{Average similarity}: Calculating the average similarity within clusters gives an insight into the coherence of the alignments, with higher values indicating more semantically consistent groupings.
  \item \textbf{The length of the aligned pair} of lines indicates whether the detected lines are within the reader's attention span and suitable for the user experience.
\end{itemize}

Comparing sentence-level and segment-level alignment, we can assume that even sentence-level alignment provides valuable insights into the differences between the translations and the original text (see Appendix D for examples in Spanish translation of \textit{I Promessi sposi}); segment-level alignment allows us to go deeper and capture more nuanced variations between the original and the translated text. For example, we can identify the omission of the end of chapter 8 in the German translation (see Table \ref{tab:religion}) and two omissions in the Russian translation of chapter 1 (see Tables \ref{tab:state}-\ref{tab:military}), which can be interpreted through the lens of cultural differences and/or censorship and could not be captured by sentence-level alignment.

\begin{longtable}{p{0.45\linewidth} p{0.45\linewidth}}
  \caption{Italian / German segment-level alignment} \label{tab:religion} \\
  \toprule
  \textbf{Original} & \textbf{German 1880} \\
  \midrule
  \endfirsthead
  \caption{(continued)} \\
  \toprule
  \textbf{Original} & \textbf{German 1880} \\
  \midrule
  \endhead
  \bottomrule
  \endfoot
  Presto, io spero, potrete ritornar sicuri a casa vostra; & Ich hoffe, ihr werdet bald ohne Gefahr in euer Haus zurückkehren können; \\
  
  a ogni modo, Dio vi provvederà, per il vostro meglio; & in jedem Falle wird Gott Alles zu euerm Besten lenken. \\
  
  e io certo mi studierò di non mancare alla grazia che mi fa, scegliendomi per suo ministro, nel servizio di voi suoi poveri cari tribolati. & \\
  
  Voi,» continuò volgendosi alle due donne, «potrete fermarvi a ***. & Und ihr», fuhr er, zu den beiden Frauen gewandt, fort, «ihr könnt euch in *** so lange aufhalten. \\
\end{longtable}

\begin{longtable}{p{0.45\linewidth} p{0.45\linewidth}}
  \caption{Italian / Russian segment-level alignment} \label{tab:military}\\
  \toprule
  \textbf{Original} & \textbf{Russian 1854} \\
  \midrule
  \endfirsthead
  \caption{(continued)} \\
  \toprule
  \textbf{Original} & \textbf{Russian 1854} \\
  \midrule
  \endhead
  \bottomrule
  \endfoot
  Ai tempi in cui accaddero i fatti che prendiamo a raccontare, quel borgo, già considerabile, era anche un castello, & Во время тех событий, которые мы намерены описать, Лекко было уже значительным местечком и маленькой крепостцей; \\
  e aveva perciò l’onore d’alloggiare un comandante, e il vantaggio di possedere una stabile guarnigione di soldati spagnoli, & вследствие чего в нем жили комендант и постоянный гарнизон испанских солдат, \\
  che insegnavan la modestia alle fanciulle e alle donne del paese, accarezzavan di tempo in tempo le spalle a qualche marito, a qualche padre; e, sul finir dell’estate, & \\
  non mancavan mai di spandersi nelle vigne, per diradar l’uve, e alleggerire a’ contadini le fatiche della vendemmia. & которые занимались собиранием винограда. \\
\end{longtable}

\begin{longtable}{p{0.45\linewidth} p{0.45\linewidth}}
  \caption{Italian / Russian segment-level alignment} \label{tab:state} \\
  \toprule
  \textbf{Original} & \textbf{Russian 1854} \\
  \midrule
  \endfirsthead
  \caption{(continued)} \\
  \toprule
  \textbf{Original} & \textbf{Russian 1854} \\
  \midrule
  \endhead
  \bottomrule
  \endfoot
  Con tutto ciò, & Несмотря на все, \\
  anzi in gran parte a cagion di ciò, & и может быть, потому именно, \\
  quelle gride, ripubblicate e rinforzate di governo in governo, non servivano ad altro che ad attestare ampollosamente l’impotenza de’ loro autori, & \\
  o, se producevan qualche effetto immediato...& если декреты имели минутную действительность...\\
\end{longtable}

\begin{figure}
  \centering
  \includegraphics[width=0.9\linewidth]{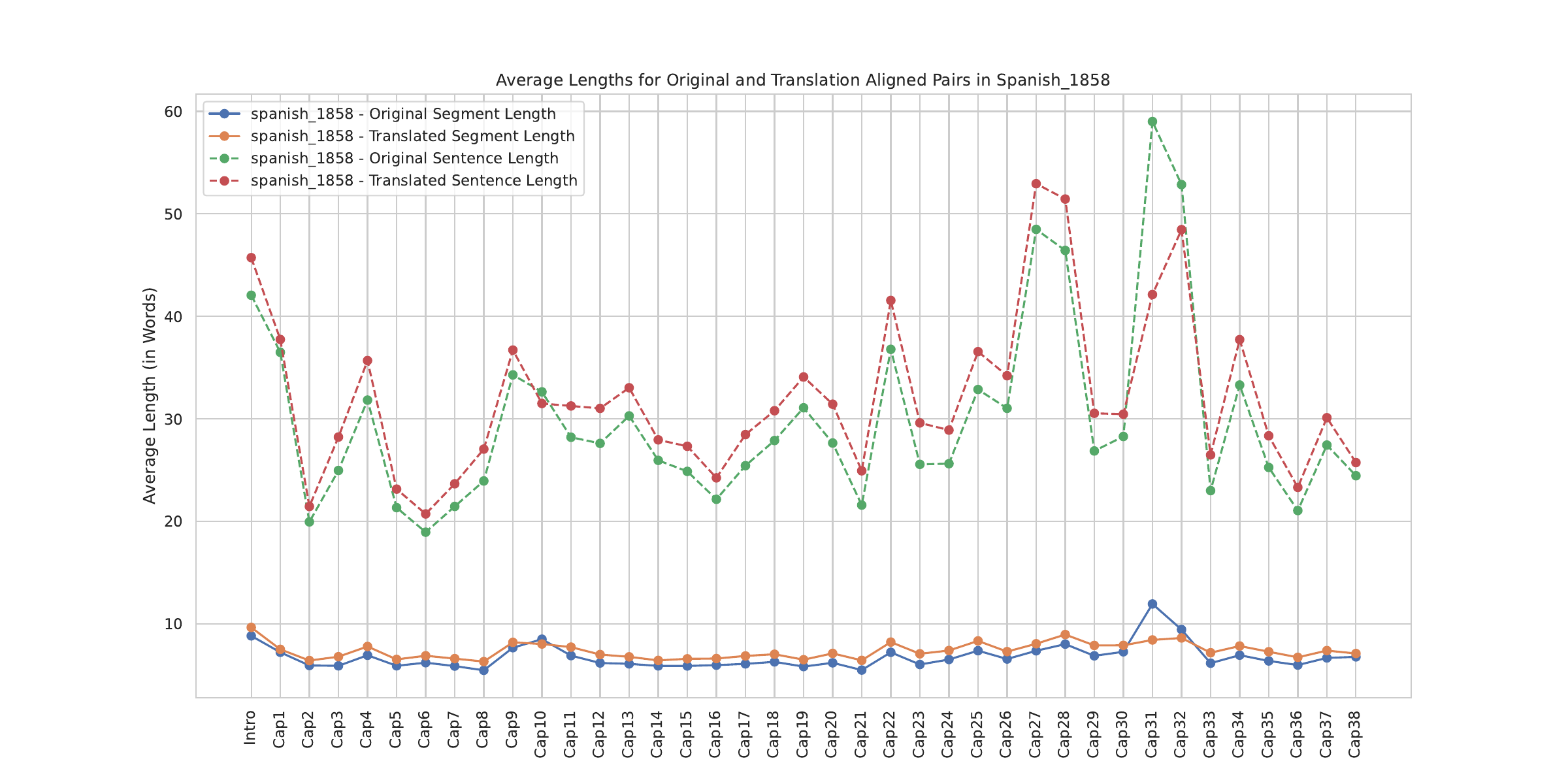}
  \caption{Reducing the Lengths of Aligned Pairs for the Spanish 1858}
  \label{fig:length}
\end{figure}

By providing a more granular and accurate alignment, the segment-level approach also allows the length of aligned pairs to be reduced (increasing their number) and makes the MDE more suitable for reader reception compared to sentence-level alignment (see Figure \ref{fig:length}).

\section{Multilingual Digital Edition Pipeline}

\begin{figure}[htbp]
  \centering
  \includegraphics[width=0.5\linewidth]{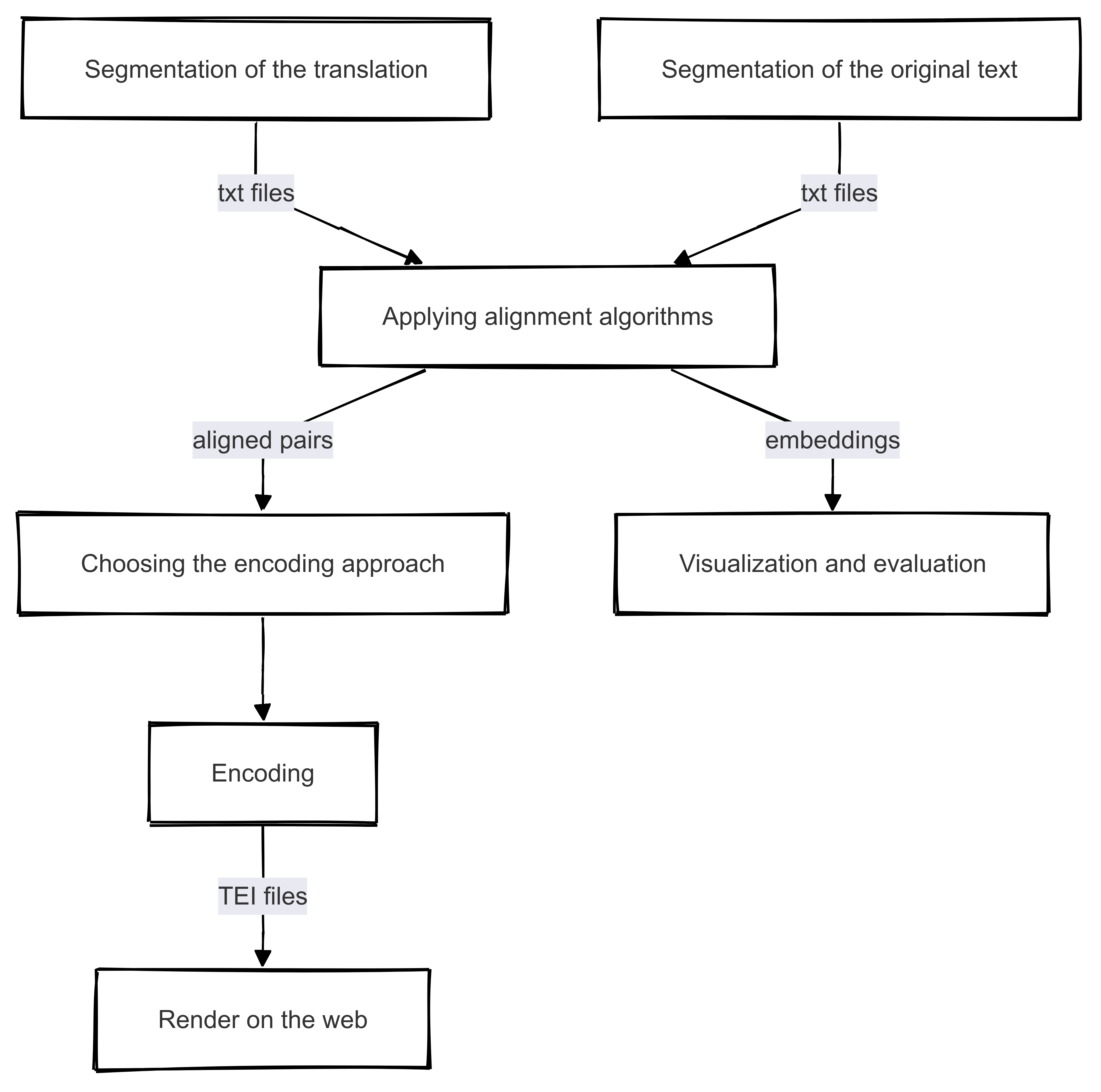}
  \caption{The Translation Alignment Pipeline}
  \label{fig:6}
\end{figure}

The automated pipeline for the MDE is proposed as a means of enabling creators to prepare annotated TEI files that are accessible, adaptable, correct, easily parsed by computational tools, and rendered for readers (see Figure \ref{fig:6}).

We start with the raw texts of the translations in TXT format, obtained after OCR and error checking. For Manzoni's text, we used TEI files with identifiers assigned to each token. This preparation allows us to take into account the irregular segmentation to be expected due to inconsistencies across the languages. 

\textbf{Step 1. The choice of the segmentation method}. Based on the above analysis and the specifics of the texts to be published, the MDE developers can select the segmentation methodology in accordance with the projected audience and the project's objectives, enabling alignment at the sentence, phrase, or word level, or a combination, giving readers the flexibility to choose their preferred option.

\textbf{Step 2-3. Segmentation of the original text and the translations}. Depending on the decision made in the first step, the text can be split into sentences, segments, or even tokens.

\textbf{Step 4. Applying alignment algorithms}: By default, we applied Bertalign with the LaBSE model, trained on 109 languages \cite{labse} to the segments obtained at the previous steps. Other multilingual sentence-transformer models, such as BGE M3-Embedding, can also be used \cite{bge-m3}. 

\textbf{Step 5. Choosing the encoding approach}. The encoding approach determines the flexibility of the alignment description for future rendering and for establishing a link between the original and translated texts. For structured texts, where each segment in the original closely corresponds to an equivalent segment in the translations, it may be appropriate to mark each segment with the same identifier. Given the complexity of multilingual alignment, we have taken a different approach. The TEI encoding of the original text includes identifiers for each token, providing granular reference points. The TEI-encoded translation text is divided into segments, each referencing the start and end identifiers from the original text, allowing for flexible and accurate alignment.

\textbf{Step 6. Encoding}. By iterating over the alignment results, we assign the referencing start and end identifiers from the original text to each aligned segment from the translation and generate a new TEI file for the translation, ensuring that all segments are accurately linked to the corresponding elements.

\textbf{Step 7. Rendering Aligned Texts on the Web}. Render the original and translated texts from the TEI files as two columns on a web page with separate XSLT templates for the original and translated text. This interactive interface allows users to click on the original text and see the corresponding translation fragment highlighted, enhancing the user experience by providing an intuitive way to explore and compare the translations side by side.

\textbf{Step 8. Visualization and evaluation}. While the highly unstable text versions\cite{medieval} or line-level aligned translations\cite{othello} can be effectively visualized with the Sankey diagram or bipartite graph, the alignment results for the modern translations can be visualized with the approach described above, based on the embedding vectors with t-SNE and clustering with DBSCAN. As for the presentation in the user interface, ideally, all multilingual translations should be comparable and aligned with each other, allowing the user to see and interpret the differences. 

\section{Future Development and Challenges}
\begin{itemize}
    \item Current alignment algorithms face challenges in accurately aligning segments with reordered content. Future work will focus on improving the alignment performance in such cases, ensuring more precise matches even when the original and translated texts differ significantly in structure.
\item Previous studies on user behavior in digital editions have analyzed log files to understand interaction patterns \cite{editing}. To gain deeper insights, we are using advanced tools such as ReactFlow to study more comprehensively how users interact with different elements of MDEs. For example, when readers view two lines in different languages side by side, the optimal reading span may differ from traditional reading practices. By analysing user interactions, we aim to determine the most effective segment length for the MDEs.
\end{itemize}

\section{Conclusion}
The proposed pipeline aims to improve the development of Multilingual Digital Editions (MDE) by ensuring that MDE is both methodologically robust and user-centered. By prioritizing user experience and usability, the pipeline adapts existing computational methods and algorithms to the specific needs of educational and research applications. 

We have also proposed new metrics for MDEs that focus on the consistency, meaningfulness and granularity of the alignment. These metrics assess the suitability of an alignment for educational and research purposes. By ensuring that the alignment is accessible to human readers while supporting translation studies, the pipeline balances conciseness and reader engagement.

\bibliography{Levchenko-CHR2024/main}
\begin{acknowledgments}
  This research was supported by the Dipartimento di Filologia Classica e Italianistica, University of Bologna, as part of the project “Manzoni online2: manoscritti e documenti inediti, tradizione e traduzioni” (CUP J34I19003370001, project code 2017CFZFAY\_003). For more information on the \textit{Leggo Manzoni} project, visit \href{https://projects.dharc.unibo.it/leggomanzoni}{https://projects.dharc.unibo.it/leggomanzoni}.
\end{acknowledgments}
\section*{Appendixes}
\appendix{}
\section{Strategies of Text/Translation Representation in MDE}
\begin{longtable}{p{0.25\linewidth} p{0.1\linewidth} p{0.1\linewidth} p{0.3\linewidth}}
  \toprule
  \multicolumn{1}{c}{\textbf{Project Name}} & \multicolumn{1}{c}{\textbf{Alignment}} & \multicolumn{1}{c}{\textbf{Comparison}} & \multicolumn{1}{c}{\textbf{Notes}} \\
  \endfirsthead
  \caption[]{Strategies of Text/Translation Representation in Multilingual Digital Editions (continued)} \\
  \toprule
  Project Name&Alignment Type&Comparison&Notes\\
  \midrule
  \endhead
  \midrule
  \multicolumn{4}{c}{\textit{1. Separate pages for the text and the translation}} \\
  \midrule
  \href{https://www.brown.edu/Departments/Italian_Studies/dweb/}{Decameron web}& - & - & \\
  \href{https://entretenida.outofthewings.org/index.html}{La entretenida by Miguel de Cervantes}& - & - & \\
  \midrule
  \multicolumn{4}{c}{\textit{2. Same page for the text and the translation with JS switcher}} \\
  \midrule
  \href{https://furnaceandfugue.org/}{Furnace and Fugue}& - & - & \\
  \midrule
  \multicolumn{4}{c}{\textit{3. Side-by-side viewer for the text and translation}} \\
  \midrule
  \href{https://miocid.wlu.edu/index.html}{Cantar de mio Cid}& - & + & \multirow{8}{0.99\linewidth}{Editions include different modes of the text on a single page with a side-by-side view, including the translation. Typically, there is no alignment, but the text length is usually just one page, making comparison straightforward. These viewers enable comparison of two or more versions of the text.} \\
  \href{https://vangoghletters.org/vg/}{Vincent van Gogh. The Letters}& - & + & \\
  \href{http://web.archive.org/web/20070530001208/http://www.tei-c.org.uk/Projects/EETS/}{Ancrene Wisse Preface}& - & + & \\
  \href{https://codexsinaiticus.org/en/project/}{The Codex Sinaiticus Project}& - & + & \\
  \href{https://dafyddapgwilym.net/}{Dafydd ap Gwilym.net}& - & + & \\
  \href{https://dante.princeton.edu/pdp/}{Princeton Dante Project}& - & + & \\
  \href{https://editions.mml.ox.ac.uk/editions/ablassgnade6/}{Ein Sermon von Ablass und Gnade}& - & + & \\
  \href{https://edition640.makingandknowing.org/}{Secrets of Craft and Nature}& - & + & \\
  \\
  \\
  \midrule
  \multicolumn{4}{c}{\textit{4. Interlinear text/translation alignment}} \\
  \midrule
  \href{http://suprasliensis.obdurodon.org/}{Codex Suprasliensis}& lines & + & The original text in Old Church Slavonic is directly followed by its corresponding parallel Greek text. \\
  \midrule
  \multicolumn{4}{c}{\textit{5. Dynamic alignment display}} \\
  \midrule
  \href{https://ebeowulf.uky.edu}{Electronic Beowulf}& lines & + & When the special view type and option are selected, and the user hovers the mouse over a line, the translation appears in a special area. \\
  \href{http://digiphil.hu/context:alomeghal}{Kassák Lajos: The Horse Dies the Birds Fly Away}& lines & + & The page displays side-by-side views of the original text and its translations into two other languages, highlighting the corresponding translated line when the mouse hovers over the original line. \\
  \href{https://cotr.ac.uk/}{The Community of the Realm in Scotland}& sentences & + & The side-by-side viewer of the Latin text with its English translation aligns the sentences, allowing users to click on the sentence number in the original text, which automatically scrolls the other side of the page to the corresponding sentence. \\
  \href{https://tabula.libripendis.eu/}{Tabula Salomonis}& lines & + & The TEI Publisher tool allows the user to highlight corresponding parts and automatically scroll when hovering over the lines. \\
  \bottomrule
\end{longtable}

\section{Translations}

\begin{itemize}
    \item \textbf{(English 1845)}. \textit{The Betrothed Lovers: A Milanese Story of the Seventeenth Century. With the Column of Infamy.} By Alessandro Manzoni. In Three Volumes. Henry Francis C. Logan. London: Longman, Brown, Green, and Longmans, Paternoster-Row.
    
    \item \textbf{(English 1876)}. \textit{The Betrothed}, by Alessandro Manzoni. London, G. Bell and Sons, 1876.
    
    \item \textbf{(English 1983)}. Alessandro Manzoni, \textit{The Betrothed}, Bruce Penman (tr.), Penguin Random House UK. London, 1983.
    
    \item \textbf{(English 2022)}. \textit{The Betrothed}. A novel, translated and with Introduction of Michael Moore, Preface by Pulitzer Prize-Winning Author Jhumpa Lahiri, Modern Library, 2022.

    \item \textbf{(Russian 1854)}. \textit{Обрученные : Медиолан. быль XVIII [!XVII] столетия, найден. и передел. Александром Манзони} / Пер. с итал. В.С. Межевича. Ч. 1-4. Москва, 1854. 4 т.; 20. (Библиотека романов, повестей, путешествий и записок, изд. Н.Н. Улитиным; Вып. 7, т. 1-2, 6-7).

    \item \textbf{(Russian 1999)}. \textit{Обрученные [Повесть из истории Милана XVII в.]} / А. Мандзони; [Пер. с итал. под ред. Н. Георгиевской, А. Эфроса]. Москва: Терра-Книжный клуб, 1999. 
        
    \item \textbf{(Dutch 1849)}. \textit{De verloofden: eene Milanesche geschiedenis uit de zeventiende eeuw. Vol. 1.} Translated by Petrus Van Limburg Brouwer. Groningen, Van Boekeren, 1849.
    
    \item \textbf{(German 1884)}. \textit{Die Verlobten: eine Mailändischer Geschichte aus dem 17. Jahrhundert, Volume 1.} 3rd ed. Regensburg, G.J. Manz, 1884.
    
    \item \textbf{(French 1874)}. \textit{Les fiancés: histoire milanaise du XVIIe siècle / Alexandre Manzoni; traduite de l'italien par Rey Dussueil.} Paris: Charpentier, 1874.
    
    \item \textbf{(Spanish 1858)}. \textit{Los desposados: historia milanesa del siglo XVII traducida del italiano, Volume 1.} México: Imp. de Andrade y Escalante, 1858.
    
    \item \textbf{(Polish 1882)}. \textit{Narzeczeni. Powieść medyolańska z XVII stulecia ze starego rękopisu spisana i przerobiona, tłum. Maria z Siermiradzkich Obrąpalska.} Warszawa, 1882.

    \item \textbf{(Chinese 1998)}. Yuehun Fufu / (Yidali) Mengzuoni (Manzoni, A.) zhu; Zhang Shihua yi. - Nanjing: Yilin chubanshe, 1998.10.

\end{itemize}

\section{Examples of Many-to-Many Alignment Type}

\begin{longtable}{p{0.45\linewidth} p{0.45\linewidth}}
  \caption{Italian / French: 3-1 alignment type}
  \label{tab:example1} \\
  \toprule
  \textbf{Italian} & \textbf{French} \\
  \midrule
  \endfirsthead
  \caption{(continued)} \\
  \toprule
  \textbf{Italian} & \textbf{French} \\
  \midrule
  \endhead
  \bottomrule
  \endfoot
  \textsuperscript{1}Sì; ma com'è dozzinale! com'è sguaiato! com'è scorretto! \par
\textsuperscript{2}Idiotismi lombardi a iosa, frasi della lingua adoperate a sproposito, grammatica arbitraria, periodi sgangherati. \par
\textsuperscript{3}E poi, qualche eleganza spagnola seminata qua e là; e poi, ch'è peggio, ne' luoghi più terribili o più pietosi della storia, a ogni occasione d'eccitar maraviglia, o di far pensare, a tutti que' passi insomma che richiedono bensì un po' di rettorica, ma rettorica discreta, fine, di buon gusto, costui non manca mai di metterci di quella sua così fatta del proemio. &
 \textsuperscript{1}Oui; mais comme il est commun! comme il est inégal! comme il est incorrect! idiotismes lombards à foison, phrases de la langue employées à rebours, constructions arbitraires, périodes boiteuses; et puis quelques petites élégances espagnoles semées ça et là; et puis, ce qui est bien pis, dans les endroits les plus terribles ou les plus touchants de son histoire, à chaque occasion d’exciter la surprise ou de faire penser, à tous les passages enfin qui demandent, il est vrai, quelques fleurs de rhétorique, mais d’une rhétorique sobre, fine, de bon goût, ce digne homme ne manque jamais d’y mettre quelque chose dans le genre de son début.
\end{longtable}

\begin{longtable}{p{0.45\linewidth} p{0.45\linewidth}}
  \caption{Italian / German: 2-2 alignment type with the overlapping sentence boundaries}
  \label{tab:example2} \\
  \toprule
  \textbf{Italian} & \textbf{German} \\
  \midrule
  \endfirsthead
  \caption{(continued)} \\
  \toprule
  \textbf{Italian} & \textbf{German} \\
  \midrule
  \endhead
  \bottomrule
  \endfoot
  \textsuperscript{1} \textit{Né alcuno dirà questa sij imperfettione del Racconto, e defformità di questo mio rozzo Parto, a meno questo tale Critico non sij persona affatto diggiuna della Filosofia}: che quanto agl'huomini in essa versati, ben vederanno nulla mancare alla sostanza di detta Narratione. \par
\textsuperscript{2} \textbf{Imperciocché, essendo cosa evidente, e da verun negata non essere i nomi se non puri purissimi accidenti...} &
 \textsuperscript{1} \textit{Und es wird gewiß niemand sagen, dies sei ein Geschichtsfälscher und eine Entstellung dieser meiner einfältigen Erzählung, es sei denn der Tadel ein Mann, der aller Weltweisheit vollständig bar wäre.} \par \par
 \textsuperscript{2} Denn man wird bald sehen, daß in Beziehung der darin vorkommenden Personen am Wesentlichsten der besagten Erzählung nichts fehle; \textbf{zumal es eine augenfällige, von niemand gelenkte Sache ist, daß Namen bloß reine Nebensachen seien…}
\end{longtable}

\begin{longtable}{p{0.45\linewidth} p{0.45\linewidth}}
  \caption{Italian / Dutch: 2-2 alignment type with the overlapping sentence boundaries}
  \label{tab:example3} \\
  \toprule
  \textbf{Italian} & \textbf{Dutch} \\
  \midrule
  \endfirsthead
  \caption{(continued)} \\
  \toprule
  \textbf{Italian} & \textbf{German} \\
  \midrule
  \endhead
  \bottomrule
  \endfoot
  \textsuperscript{1} \textit{Però alla mia debolezza non è lecito solleuarsi a tal'argomenti, e sublimità pericolose, con aggirarsi tra Labirinti de' Politici maneggj, et il rimbombo de' bellici Oricalchi}: solo che hauendo hauuto notitia di fatti memorabili, se ben capitorno a gente meccaniche, e di piccol affare, mi accingo di lasciarne memoria a Posteri, con far di tutto schietta e genuinamente il Racconto, ouuero sia Relatione.  \par
\textsuperscript{2} Nella quale si vedrà in angusto Teatro luttuose Traggedie d'horrori, e Scene di malvaggità grandiosa, con intermezi d'Imprese virtuose e buontà angeliche, opposte alle operationi diaboliche. &
 \textsuperscript{1} \textit{Doch mijn' geringeren krachten is het niet gegeven zich tot zoo hooge vlugt, tot zulk eene gevaarvolle verhevenheid te verheffen, en zich te wagen in den doolhof der staatkundige spitsvondigheden of te midden van het geschal der schorre krijgsklaroenen. } \par \par
 \textsuperscript{2} Naardemaal 'er dus eenige merkwaardige gebeurtenissen ter mijner kennis gekomen zijn, welke, wel is waar, slechts menschen van gering bedrijve en lage geboorte betreffen, maar des alniettemin eene rijke vertooning opleveren van droevige en vreesselijke ongevallen, voorbeelden van drieste boosheid, doormengd met vrome ondernemingen en verheerljkt door het zielesterkend schouwspel van hemelsche deugd, in onophoudelijken strijd met de gruwelijke aanslagen der helle, zoo heb ik besloten mij aan te gorden om daarvan der nakomelingschap een getrouw en nauwkeurig Verhaal ofte Relaas achterlaten.
\end{longtable}

\begin{longtable}{p{0.45\linewidth} p{0.45\linewidth}}
  \caption{Italian / Spanish: 2-3 alignment type with the overlapping sentence boundaries}
  \label{tab:example4} \\
  \toprule
  \textbf{Italian} & \textbf{Spanish} \\
  \midrule
  \endfirsthead
  \caption{(continued)} \\
  \toprule
  \textbf{Italian} & \textbf{Spanish} \\
  \midrule
  \endhead
  \bottomrule
  \endfoot
  \textsuperscript{1} \textit{Ma che? quando siamo stati al punto di raccapezzar tutte le dette obiezioni e risposte, per disporle con qualche ordine, misericordia! venivano a fare un libro.}   \par
\textsuperscript{2} Veduta la qual cosa, abbiam messo da parte il pensiero, per due ragioni che il lettore troverà certamente buone: la prima, che un libro impiegato a giustificarne un altro, anzi lo stile d'un altro, potrebbe parer cosa ridicola: la seconda, che di libri basta uno per volta, quando non è d'avanzo. &
 \textsuperscript{1} \textit{Pero, ¡oh cielos! llegado el momento de recapitular las objeciones y sus respuestas y el de ordenarlas, hallamos, que habíamos hecho un libro}: visto lo cual, abandonamos nuestro intento por dos razones, que sin duda alguna el lector considerará oportunas. — \par \par
 \textsuperscript{2} La primera, porque temimos que el hacer un libro para justificar otro, ó solo su estilo, parecería cosa ridícula. \par
 \textsuperscript{3} La segunda, porque creemos que es suficiente, cuando no excesivo, el publicar un solo libro á la vez.
\end{longtable}

\section{Examples of Omission Captured through Sentence-Level Alignment}

\begin{longtable}{p{0.45\linewidth} p{0.45\linewidth}}
  \caption{Italian / Spanish segment level alignment for the Chapter 7}
  \label{tab:example5} \\
  \toprule
  \textbf{Italian} & \textbf{Spanish 1858} \\
  \midrule
  \endfirsthead
  \caption{(continued)} \\
  \toprule
  \textbf{Italian} & \textbf{Spanish 1858} \\
  \midrule
  \endhead
  \bottomrule
  \endfoot
  Gertrude domandò sommessamente e tremando, che cosa dovesse fare. & Gertrudis con mucha timidez pidió la explicación de aquellas palabras y lo que debía hacer en consecuencia. \\
  
  Il principe (non ci regge il cuore di dargli in questo momento il titolo di padre) non rispose direttamente, ma cominciò a parlare a lungo del fallo di Gertrude: e quelle parole frizzavano sull'animo della poveretta, come lo scorrere d'una mano ruvida sur una ferita. &  \\
  Continuò dicendo che, quand'anche... caso mai... che avesse avuto prima qualche intenzione di collocarla nel secolo, lei stessa ci aveva messo ora un ostacolo insuperabile; giacché a un cavalier d'onore, com'era lui, non sarebbe mai bastato l'animo di regalare a un galantuomo una signorina che aveva dato un tal saggio di sé. & Él continuó diciendo que... “á pesar de lo ocurrido... en el caso en que... hubiera sido con la intención de establecerse en el mundo, ella había contraído un lazo indisoluble y había creado un obstáculo invencible. Hombre de honor como era, jamás se habría atrevido á presentarla á ningún caballero después de tales antecedentes”. \\
  «Ebbene, non si parli più del passato: tutto è cancellato. & En hora buena; no hablemos más de lo pasado: todo está olvidado ya.\\
  Avete preso il solo partito onorevole, conveniente, che vi rimanesse; ma perché l'avete preso di buona voglia, e con buona maniera, tocca a me a farvelo riuscir gradito in tutto e per tutto: tocca a me a farne tornare tutto il vantaggio e tutto il merito sopra di voi. & \\
  Ne prendo io la cura.» & \\
\end{longtable}

\section{Examples of Omission Captured through Segment-Level Alignment}

\begin{longtable}{p{0.45\linewidth} p{0.45\linewidth}}
  \caption{Italian / Spanish segment level alignment for the Chapter 7}
  \label{tab:example6} \\
  \toprule
  \textbf{Italian} & \textbf{Spanish 1858} \\
  \midrule
  \endfirsthead
  \caption{(continued)} \\
  \toprule
  \textbf{Italian} & \textbf{Spanish 1858} \\
  \midrule
  \endhead
  \bottomrule
  \endfoot
  «Brava! bene!» esclamarono, a una voce, la madre e il figlio,&  — Muy bien, muy bien, exclamaron á la par madre é hijo.\\
e l'uno dopo l'altra abbracciaron Gertrude;& \\
la quale ricevette queste accoglienze con lacrime,&\\
che furono interpretate per lacrime di consolazione.&\\
Allora il principe si diffuse a spiegar ciò che farebbe per render lieta e splendida la sorte della figlia.&\\
Parlò delle distinzioni di cui goderebbe nel monastero e nel paese;& Entonces el príncipe habló de las distinciones que Gertrudis habría de tener en el convento y en el país.\\
che, là sarebbe come una principessa, come la rappresentante della famiglia; che, appena l'età l'avrebbe permesso, sarebbe innalzata alla prima dignità; e, intanto, non sarebbe soggetta che di nome.&\\
\end{longtable}

\end{document}